\documentclass[10pt,twocolumn,letterpaper]{article}

\usepackage{wacv}
\usepackage{times}
\usepackage{epsfig}
\usepackage{graphicx}
\usepackage{amsmath}
\usepackage{amssymb}
\usepackage{lipsum}
\usepackage{flushend}

\def\wacvPaperID{263} 

\wacvfinalcopy 

\ifwacvfinal
\def\assignedStartPage{1} 
\fi

\ifwacvfinal
\usepackage[breaklinks=true,bookmarks=false]{hyperref}
\else
\usepackage[pagebackref=true,breaklinks=true,colorlinks,bookmarks=false]{hyperref}
\fi

\ifwacvfinal
\else
\fi

\begin{document}

\title{Keypoint-Aligned Embeddings for Image Retrieval and Re-identification}

\author{Olga Moskvyak, Frederic Maire, Feras Dayoub\\
Queensland University of Technology\\
{\tt\small \{o.moskvyak, f.maire, feras.dayoub\}@qut.edu.au}
\and
Mahsa Baktashmotlagh\\
University of Queensland\\
{\tt\small m.baktashmotlagh@uq.edu.au}
}

\maketitle

\begin{abstract}
   Learning embeddings that are invariant to the pose of the object is crucial in visual image retrieval and re-identification.
   The existing approaches for person, vehicle, or animal re-identification tasks suffer from high intra-class variance due to deformable shapes and different camera viewpoints.
   To overcome this limitation, we propose to align the image embedding with a predefined order of the keypoints.
   The proposed keypoint aligned embeddings model (KAE-Net) learns part-level features via multi-task learning which is guided by keypoint locations. 
   More specifically, KAE-Net extracts channels from a feature map activated by a specific keypoint through learning the auxiliary task of heatmap reconstruction for this keypoint.
   The KAE-Net is compact, generic and conceptually simple. 
   It achieves state of the art performance on the benchmark datasets of CUB-200-2011, Cars196 and VeRi-776 for retrieval and re-identification tasks.

\end{abstract}


\section{Introduction}

Learning pose invariant image embeddings is critical for visual search tasks such as image retrieval, person or vehicle identification. 
Pose variations means that the position of parts of the class instance (a person or a car) within the image is not known and the poses of the objects across the dataset are not aligned. 
As an example, the same car looks differently from the front and the rear and the appearance of a bird changes dramatically with its activities (e.g., flying or sitting). 
Large intra-class variations caused by a variety of viewing angles pose a challenge for many image retrieval and identification tasks when learning discriminative embeddings.

In recent years, this problem has been approached by learning an image representation in such a way that images from the same class (e.g., a bird species, a person or a car) are mapped closer to each other compared to the images from different classes. 
To this end, researchers have proposed new loss functions~\cite{res-ms-loss, person-part-loss, recall-metric, res-soft-triplet}, introduced new methods to mine useful training samples \cite{res-sampling, res-easytri} and developed approaches to reduce embedding space diversity \cite{dml-survey, res-horde}. 
These methods rely on a large corpus of training data to learn the relevant features and depend on the discriminating power of the samples that are presented to the network.

Another stream of research is focused on image alignment as an effective method to accommodate to viewpoint variations and it is used successfully in face re-identification thanks to a range of robust and accurate methods to detect facial landmarks \cite{facial-ldm-detection}. 
However, this approach is limited in a sense that it requires precise detection of keypoints and in case of non-overlapping views (e.g., a front and a back view of a car) image alignment is not possible.

\begin{figure}[t]
\begin{center}
   \includegraphics[width=0.8\linewidth]{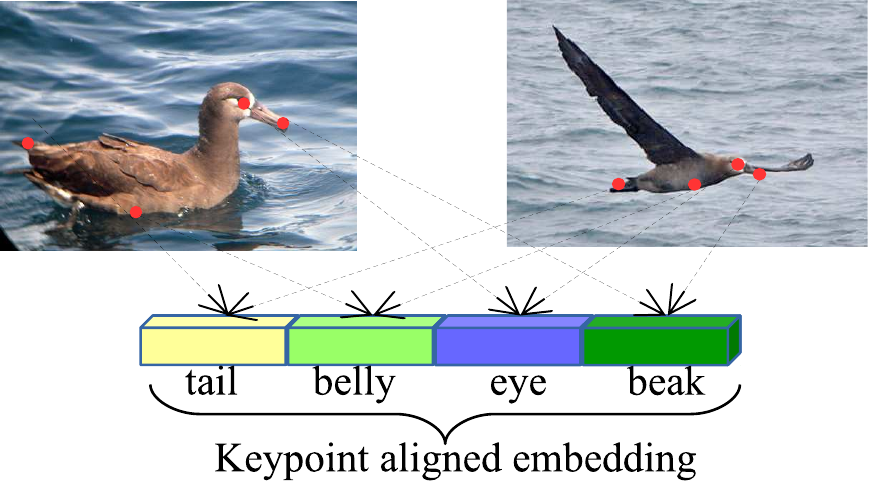}
\end{center}
   \caption{Our KAE-Net learns an image embedding that consists of several subvectors where each subvector corresponds to a specific keypoint. The figure shows a keypoint aligned embedding with subvectors for four bird keypoints: beak, right eye, belly, tail.}
\label{fig:intro-fig}
\end{figure}

To overcome this limitation, instead of aligning images, we propose to achieve pose invariance by learning a pose invariant representation. Our work approaches metric learning from a different angle: We learn embeddings that are pose invariant by training a neural network on the auxiliary task of predicting the locations of keypoints.

More specifically, we learn a \textit{keypoint aligned embedding} that consists of subvectors where each subvector encodes a specific keypoint (Figure~\ref{fig:intro-fig}). This is motivated by the fact that the average pooling across spatial dimensions in the feature map of a convolutional neural network (CNN) generates a saliency map that depicts the most distinctive parts of the image \cite{review-saliency}. Moreover, different channels in the feature map are activated more strongly at those regions which correspond to specific parts of the object in the image (e.g., person's body parts) \cite{person-part-loss, review-occluded}. We utilize the correspondence between channels of the feature map and image parts to align an image embedding with keypoints.
In other words, an image embedding consists of a number of subvectors where each one corresponds to a specific part of the object as shown in Figure~\ref{fig:intro-fig}.

Our proposed \textit{keypoint aligned embeddings model} (KAE-Net) explicitly learns subvectors representing each keypoint in a multi-task fashion guided by the auxiliary task of keypoint reconstruction.  Multi-task learning allows us to jointly optimize both localization of keypoints and learning discriminative image embeddings. 
Our proposed system can be seen as an improvement on existing works that attempt to localise salient parts and learn the class label separately \cite{review-birds-aligned-object, review-person-pose-invariant}, generate discriminative region proposals in parallel with feature learning \cite{roi-part-detection}, or utilize unnecessarily complicated part attention models \cite{review-object-part}.

Our contribution is three-fold:
\begin{itemize}
    \item We propose KAE-Net, a conceptually simple, but effective model to learn pose invariant image embeddings;
    \item We demonstrate that it is possible to learn pose invariant embeddings with the auxiliary task of reconstructing pose information;
    \item Our approach is not domain specific and achieves significant improvement over the state-of-the-art on benchmarks CUB-200-2011 \cite{cub-200-2011}, Cars196 \cite{cars196} and VeRi-776 \cite{veri-1, veri-2, veri-3}.

\end{itemize}

The paper is organised as follows.
Related work on embedding learning using pose information and multi-task learning is reviewed in Section~\ref{sec:related-work}.
KAE-Net architecture is described in Section~\ref{sec:methodology}.
Experimental settings, datasets and results are discussed in Section~\ref{sec:experiments}.
       
\section{Related work}
\label{sec:related-work}

\begin{figure*}[t]
\begin{center}
   \includegraphics[width=0.9\linewidth]{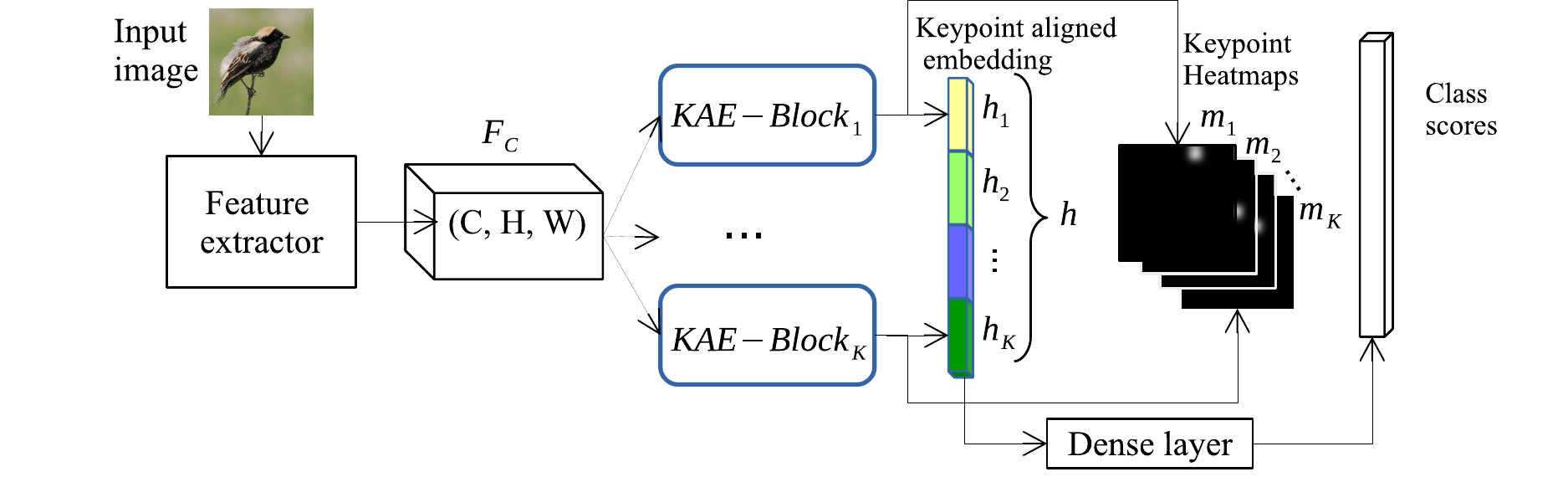}
\end{center}
   \caption{KAE-Net architecture. KAE-Blocks do not share weights. The number of KAE-Blocks is equal to the number of keypoints defined for the object of interest. KAE-Net outputs a keypoint aligned embedding, class scores and keypoint heatmaps.}
\label{fig:kae-net}
\end{figure*}

In this section, we discuss methods that use some degree of pose information (object parts, segmentation masks, or keypoints) to facilitate embedding learning.
We also cover multi-task learning approaches for image retrieval and re-identification that are related to our work.

\subsection{Utilizing Pose Information}

Pose information has been used for vehicle re-identification which is a task of identifying a specific vehicle from the images taken by several non-overlapping cameras on the street. 
The most successful methods for vehicle re-identification involve deep metric learning approaches that exploit keypoint locations \cite{vehicle-survey, vehicle-triplet-baseline, vehicle-orientation-invariant-2017, vehicle-pamtri, vehicle-viewpoint-aware}.
Wang~et~al.~\cite{vehicle-orientation-invariant-2017} proposed an orientation invariant feature embedding module that learns viewpoint invariant features based on 20 keypoint locations. PAMTRI (Pose-Aware Multi-Task Re-Identification) framework \cite{vehicle-pamtri} explicitly receives information about a vehicle pose via keypoints, heatmaps, and segments as additional inputs.
Chu~et~al.~\cite{vehicle-viewpoint-aware} propose a view-point aware network that learns two separate metrics for similar and different viewpoints in two feature spaces.


Including pose information can be also beneficial for a person re-identification task where a person should be re-identified across different cameras. Several works \cite{review-person-pose-driven, review-person-pose-invariant} explicitly leverage cues about body part locations and learn feature representations from the full image and different local parts. 
\cite{person-part-loss, person-coarse} use feature map responses to localize object parts without supervision, and pool over region proposals in a weakly supervised manner. 
Information about the 2D pose in the form of a heatmap is added explicitly to the image input as additional channels in \cite{person-pose-sensitive, Moskvyak2020LearningLG}. 
However, it is unclear how this additional input is used by the model to learn an embedding. To overcome this limitation, \cite{review-birds-aligned-object} utilizes keypoint locations and crops patches around the keypoints, extracts features, and concatenates them in one embedding vector for prediction. This method has been used for fine-grained bird classification.

\subsection{Multi-task Learning}

Multi-task learning is used to simultaneously optimize for several tasks and improve the generalization with the supervision from several objectives.
Apart from the main tasks, which are the final output of the model, auxiliary tasks serve as additional regularization for learning a rich representation of an image \cite{Liu2019AuxiliaryLF}.
Auxiliary modules are usually removed during the inference stage.

Learning an auxiliary task improves performance in scene understanding \cite{Liebel2018AuxiliaryTI, Liu2019AuxiliaryLF}, and vehicle re-identification \cite{vehicle-pamtri}.
Ding~et~al.~\cite{person-coarse} show the benefits of multi-task attention for learning a part-aware person representation.

Our work is close to  \cite{review-birds-aligned-object} in spirit, as it leverages keypoint locations. However, in \cite{review-birds-aligned-object}, cropping patches removes the context and training a separate network for each keypoint patch results in a large and complicated network to train. To avoid this problem, our method guides the learning of a keypoint aligned embedding using the auxiliary task of keypoint reconstruction.
Jointly learning an auxiliary task boosts the performance of the main task \cite{Liebel2018AuxiliaryTI}. 

Different from \cite{person-part-loss, person-coarse}, we do not pool over region proposals and use full spatial dimensions of the feature map for pooling. Our approach uses guidance from keypoint heatmaps to select channels in the feature maps that are the most responsive for the specific keypoint.
    
\section{Proposed Approach}
\label{sec:methodology}

In this section, we introduce a \textit{keypoint embedding block} (KAE-Block), a new sub-network for learning pose invariant keypoint embeddings, and explain the architecture of the whole network.
Our network learns features corresponding to each keypoint to construct a representation that is aligned with the order of the keypoints.
The order of the keypoints is arbitrary and is specified with the model definition.

Note that since our method is generic, and is applicable to other-than-human such as vehicles and birds, we use the term  \textit{keypoints} instead of body joints or body parts.

\subsection{Keypoint Aligned Embedding Network}

The backbone of KAE-Net is a CNN that outputs a feature map $F_C$ of size $C\times H \times W$.
It has been observed in \cite{review-saliency} that different channels in the feature map extract various meaningful parts of the images.
Building on the idea that only a subset of the channels in the feature map is activated for each keypoint, we train the KAE-Block to select these channels. 
To this end, we jointly optimize for the main task of embedding learning and the auxiliary task of keypoint heatmap reconstruction.
We describe KAE-Block in detail in Section~\ref{sec:kae-block}.

The model assigns one KAE-Block to each keypoint (Figure~\ref{fig:kae-net}). 
The user-defined order of the keypoints is fixed for the training and inference.
During training, the model outputs a keypoint aligned embedding, class scores, and keypoint heatmaps for each input image. 
During inference, only the image embedding is computed.

\subsection{KAE-Block}
\label{sec:kae-block}

\begin{figure}[t]
\begin{center}
   \includegraphics[width=0.8\linewidth]{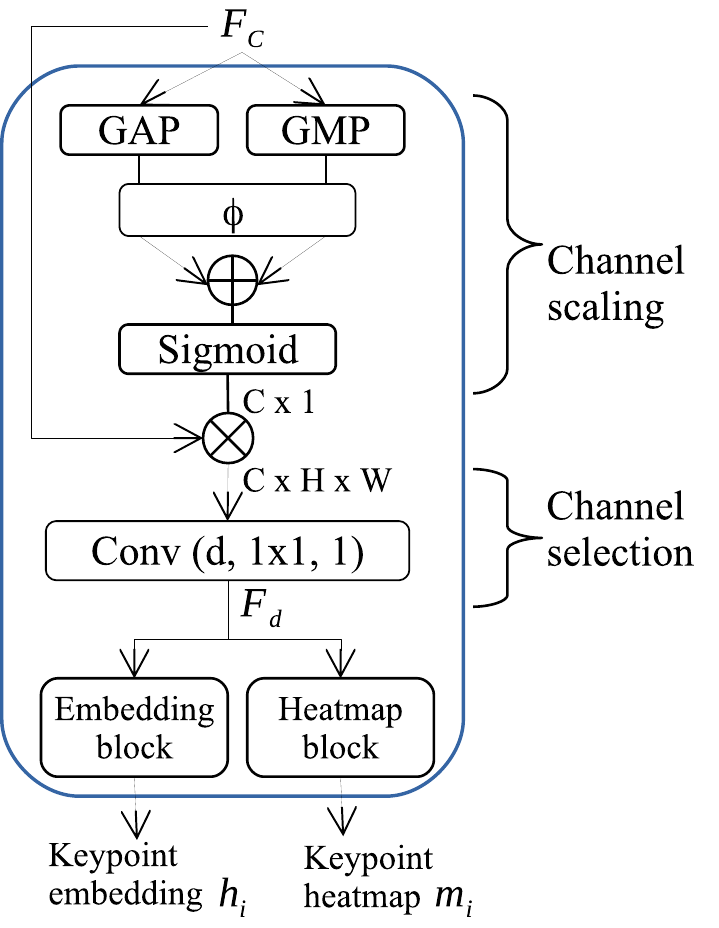}
\end{center}
   \caption{A KAE-Block receives a feature map $F_C$ of the size $C \times H \times W$ as an input and outputs a keypoint embedding $h_i$ and a keypoint heatmap $m_i$. A KAE-Block consists of a channel rescaling part, a channel selection part, an embedding block and a heatmap reconstruction block.}
\label{fig:kae-block}
\end{figure}

A KAE-Block consists of several parts: channel rescaling, channel selection, an embedding block, and a heatmap reconstruction block (Figure~\ref{fig:kae-block}).
A KAE-Block accepts a feature map $F_C$ of the size $C \times H \times W$ and outputs an embedding vector of the length $d=C/r$, with $r$ a reduction rate, and a reconstructed keypoint heatmap of the size $1 \times H_{hm} \times W_{hm}$, with $H_{hm} \geq H$, and $W_{hm} \geq W$.

Global average pooling (GAP) over feature maps \cite{lin2013network} is commonly adopted to aggregate spatial information.
Global max pooling (GMP) also provides an important cue about distinctive object features.
We use both GAP and GMP as it was proved by Woo~et~al.~\cite{Woo2018CBAMCB} that using both descriptors is effective for learning channel attention.

A channel rescaling part of the KAE-Block first pools the feature map $F_C$ using GAP and GMP into the vectors $F_C^a$ and $F_C^m$, respectively.
Both vectors are passed to a small shared network $\phi$. 
Then, the output vectors $\phi(F_C^a)$ and  $\phi(F_C^m)$ are summed elementwise followed by the sigmoid function $\sigma$ to normalize the values between 0 and 1.
The produced channel attention map $M_C$ is essentially a vector of weights to scale the original features maps $F_C$:

\begin{equation}
    M_C = \sigma(\phi(F_C^a) \oplus \phi(F_C^m))
\end{equation}

Inspired by~\cite{attention-harmonious}, the shared network $\phi$ is a fully connected network with $d$ hidden neurons and $C$ output neurons. Each layer is followed by ReLU and Batch Normalization \cite{Ioffe2015BatchNA}.
We use $d$ neurons in the hidden layer to reduce the number of parameters.

A KAE-Block scales feature maps $F_C$ with the attention vector $M_C$ by element-wise multiplication.
The output $F_C \odot M_C$ is fed into $1\times1$ convolution with $d$ filters to get a feature map $F_{d}$ with the shape $ d \times H \times W$.
Note that, while the number of channels is reduced, the spatial dimension remains the same.

Learning channel scaling and selection is supervised by two tasks: learning an embedding and reconstructing a keypoint heatmap.
A keypoint reconstruction part of a KAE-Block learns a heatmap from the reduced feature map $F_{d}$ using a series of deconvolutional layers. 
An embedding learning part applies global max pooling to $F_{d}$ followed by a fully connected layer to output a keypoint embedding.

The model uses one separate KAE-Block for each keypoint:
\begin{equation}
    (h_i, m_i) =  \text{KAE-Block}_{i} (F_C)
\end{equation}
for $ i \in \{1, ..., K\}$, with $K$ the number of keypoints, $h_i$ a keypoint embedding, and $m_i$ an output heatmap for the $i$-th keypoint.

Ultimately, the $K$ keypoint embeddings are concatenated into one keypoint aligned embedding:
\begin{equation}
    h = [h_{1}, h_2, ..., h_{K}]. 
\end{equation}

\subsection{Loss Function}

The KAE-Net learns keypoint aligned embeddings with a combination of losses which are jointly optimized for the main and the auxiliary tasks.
More specifically, the overall objective function optimized during the training is the weighted sum of the following losses:

\begin{equation}
    L = \lambda_{k} L_{\text{kp-tri}} + \lambda_{h} L_{\text{hm}} + \lambda_{v} L_{\text{vis}} + \lambda_{c} L_{\text{ce}},
\end{equation}
We select the weights $\lambda$ such that all the losses remain at the same scale. 
We now discuss each component of the loss function.

\textbf{$L_{\text{kp-tri}}$}: We compute a triplet loss \cite{batch-hard} $L_{\text{kp-tri}}$ for the image embedding $h$ as well as for subvectors $[h_{1}, h_2, ..., h_{K}]$.
The triplet loss  aims to learn a representation where the distance between the points in the embedding space of intra-class image pairs will be smaller than the one of the inter-class image pairs.

Some keypoints may be not visible in the image so computing loss over embeddings $h_i$ for these keypoints would result in learning with a lot of noise. Thus, the triplet loss $L_{\text{tri}}$ for each keypoint embedding is computed only on visible keypoints.

We use BatchHard \cite{batch-hard} triplet mining strategy that selects the hardest negative and positive pairs within a batch.
To prevent the noisy embeddings from contributing as hard examples to the loss, we disregard the embeddings of non-visible keypoints. We further compute the triplet loss over the whole image embedding $h$ to make it robust to noise: 

\begin{equation}
    \begin{aligned}
        L_{\text{kp-tri}} &= \frac{1}{K} \sum_{i=1}^{K} \sum_{(h_i^a, h_i^p, h_i^n) \in V_i} L_{\text{tri}}(h_i^a, h_i^p, h_i^n)  \\
        & + \sum_{} L_{\text{tri}} (h^a, h^p, h^n),
    \end{aligned}
\end{equation}

Here, $(h_i^a, h_i^p, h_i^n)$ are triplets of embeddings in a batch that belong to the subset $V_i$ of embeddings, with the $i$-th keypoint visible, and $(h^a, h^p, h^n)$ denote triplets of image embeddings.

\textbf{$L_{\text{hm}}$}: The reconstruction of keypoint heatmaps is supervised with the mean squared error loss  $L_{\text{hm}}$ between an on output and ground truth heatmaps. 

\textbf{$L_{\text{vis}}$}: In addition to the reconstruction loss, we guide the learning of the heatmaps for not-visible keypoints with the  binary cross-entropy loss. In particular, if the keypoint is not visible, the corresponding ground truth heatmap is all zeros. Therefore, to suppress the signal for  not-visible keypoints, we apply binary cross entropy $L_{\text{vis}}$ on max pooled reconstructed and ground truth heatmaps:
\begin{equation}
    \begin{aligned}
    L_{\text{vis}} &=  -\frac{1}{K N} \sum_{j=1}^{N} \sum_{i=1}^{K} \Big( y_{ij} \cdot \log \sigma(\max(m_{ij})) \\
        &+ (1 - y_{ij}) \cdot \log \big(1 - \sigma(\max(m_{ij}))\big) \Big)
    \end{aligned}    
\end{equation}
with $m_{ij}$ an output heatmap and $K$ the number of keypoints. 

Here, $y_{ij}$ is a ground truth visibility for the $i$-th keypoint of $j$-th example in a batch of size $N$. A ground truth visibility $y_{ij}$ is the maximum of the corresponding ground truth heatmap which is zero if the keypoint is not visible and one otherwise. 
Function $\sigma$ is a sigmoid function that is combined with the $\log$ in the loss instead of applied as a layer for numerical stability.

\textbf{$L_{\text{ce}}$}: As proven to be effective on re-identification baselines \cite{strong-baseline-personreid, Wojke2018DeepCM}, we apply cross-entropy loss $L_{\text{ce}}$ for training class scores.

\begin{table*}[t]
\begin{center}
\begin{tabular}{ll|cccc|cccc}
\multicolumn{1}{c}{} & \multicolumn{1}{c|}{Backbone} & \multicolumn{4}{c|}{CUB-200-2011} & \multicolumn{4}{c}{Cars196} \\ \cline{3-10} 
\multicolumn{1}{c}{Method} & \multicolumn{1}{c|}{model} & R@1 & R@2 & R@4 & R@8 & R@1 & R@2 & R@4 & R@8 \\ \hline
Margin \cite{res-sampling} & ResNet50 & 63.6 & 74.4 & 83.1 & 90.6 & 79.6 & 86.5 & 91.9 & 95.1 \\
EPSHN512 \cite{res-easytri} & ResNet50 & 64.9 & 75.3 & 83.5 & - & 82.7 & 89.3 & 93.0 & - \\
NormSoftmax \cite{res-strbase} & ResNet50 & 65.3 & 76.7 & 85.4 & 91.8 & 89.3 & 94.1 & 96.4 & 98.0 \\
SoftTriple \cite{res-soft-triplet} & BN-Inception & 65.4 & 76.4 & 84.5 & 90.4 & 84.5 & 90.7 & 94.5 & 96.9 \\
MS512 \cite{res-ms-loss} & BN-Inception & 65.7 & 77.0 & 86.3 & 91.2 & 84.1 & 90.4 & 94.0 & 96.5 \\
ABE+HORDE \cite{res-horde} & BN-Inception & 66.8 & 77.4 & 85.1 & 91.0 & 86.2 & 91.9 & 95.1 & 97.2 \\ \hline
KAE-Net (ours) & ResNet50 & \textbf{74.2} & \textbf{83.3} & \textbf{89.1} & \textbf{93.2} & \textbf{91.1}  & \textbf{94.9} & \textbf{96.9} & \textbf{98.1} \\
\end{tabular}
\end{center}
\caption{Comparison of retrieval performance with recent methods on CUB-200-2011 and Cars196. Results in percents.}
\label{tab:cub-cars-results}
\end{table*}

\subsection{Evaluation}
We evaluate our model on the tasks of image retrieval and re-identification, following the metrics adopted in the literature for these tasks.

For retrieval, the dataset is split into training and testing subsets with disjoint classes. 
The retrieval metric Recall@R  is the percentage of the testing examples whose set of $R$ nearest neighbours includes at least one example of the same class \cite{recall-metric}.

For the re-identification task, the adopted evaluation schema assumes two datasets for evaluation: a query and a gallery sets with identities not present during training.
For multi-camera scenario, each example is labelled with a camera ID in addition to the identity ID.
For each query image the predictions are selected from the gallery set excluding the images that share the same identity and camera ID as the query. 
Evaluation metrics adopted for this dataset are Cumulative Match Curve for top 1 (CMC@1) and top 5 (CMC@5) matches,  as well as the mean Average Precision (mAP) \cite{veri-1}.

Many retrieval and re-identification pipelines apply re-ranking step at the end to rearrange the predictions and assign higher rank to some of the samples, which helps with improving their performance. Note that, we do not adopt these techniques, confirming that the superiority of our approach is independent of any re-ranking strategies.
               
\section{Experiments}
\label{sec:experiments}

In this section, we compare the results of our KAE-Net with the state-of-the-art on three benchmark datasets of CUB-200-2011, Cars196 and VeRi-776 for image retrieval. 
The reason behind choosing these datasets is the availability of image level labels and keypoint annotations which are required in our approach. 
Note that, our method is applicable to the datasets with the same type of objects categories (e.g.~cars, birds) where keypoints are consistent across images.

\textbf{Implementation details:} We use ResNet50 \cite{resnet} as the backbone network that outputs feature maps with the number of channels $C=2048$. Input images in all our experiments are resized to $256 \times 256$, resulting in feature maps of size $2048 \times 16 \times 16$. The training images are augmented by random horizontal flip and normalization.

We set the reduction rate to $r=32$, meaning that features are reduced to the size $64 \times 16 \times 16$ for each keypoint. To get larger spatial feature maps before layer 4, we remove the dimensionality reduction.

Our model is learning to reconstruct heatmaps that are created by placing a Gaussian kernel with variance 1 in the keypoint coordinate.
We use heatmaps of size $64 \times 64$ which are four times smaller than the input image. If a keypoint is not visible then it is represented with a heatmap with all zeros. Keypoint coordinates are only used during training and are not required for inference.

Euclidean distance (L2) is used to compute similarity score between query and gallery images during training and testing.
The training batch consists of 64 images with 8 images per class/identity.


KAE-Net is trained in several steps.
In the first step, we finetune the backbone ResNet50 pretrained on ImageNet \cite{resnet} on the training set by optimizing the sum of triplet and cross-entropy losses with the learning rate of $1 \times 10^{-4}$.
Then, we freeze the backbone weights and train randomly initialised weights of KAE-Blocks with the target loss $L$ with the learning rate of $1 \times 10^{-3}$. Finally, we tune the whole model with the loss $L$ and the learning rate of $1 \times 10^{-4}$.

We adopt Adam optimizer with a weight decay of $1 \times 10^{-4}$ to train KAE-Net.
We use the triplet loss with a soft margin \cite{batch-hard} to avoid margin parameters.
The weights of loss components are selected to keep all losses at the same scale:
$\lambda_{k}=10$, $\lambda_{h}=1000$, $\lambda_{v}=1$, $\lambda_{c}=1$.

\subsection{Datasets}
\label{sec:datasets}

\textbf{CUB-200-2011}

The dataset CUB-200-2011 \cite{cub-200-2011} of bird images is annotated with fine-grained bird category labels and 15 body parts such as beak, right eye, forehead, left eye, back, breast.
We use the split adopted for the image retrieval task with the first 100 classes selected for training and the rest 100 classes for testing. 
Each class has on around 40-60 images.
The adopted evaluation metric for retrieval on CUB-200-2011 is Recall@1, Recall@2, Recall@4 and Recall@8.

\textbf{VeRi-776}

VeRi-776 \cite{veri-1, veri-2, veri-3} is an image dataset of 776 distinct vehicles that were taken with 20 non-overlapping cameras in variety of orientations including front, rear and side views.
The licence plates are erased from the images.
The dataset has 49,357 images in total with 37,778 (576 identities) for training and 11,579 (200 identities) for testing. 
The query set consists of 1,678 images selected from the query set.
The adopted evaluation metric for VeRi is CMC@1, CMC@5 and mAP in cross-camera setting \cite{veri-1}.

We use the ground truth annotations for 20 keypoints on the training set publicly provided in \cite{vehicle-orientation-invariant-2017}. 

\textbf{Cars196}

Cars196 \cite{cars196} is a popular benchmark for image retrieval with 16,185 images of 196 car models. 
We use the conventional protocol of splitting and use the first 98 classes (8,054 images) for training and the remaining 98 classes (8,131 images) for testing.
The evaluation metric is the same as for CUB-200-2011.

Cars196 dataset does not have keypoint annotations.
We fine-tuned a pose estimation network HRNet \cite{sun2019deep, xiao2018simple} from a provided checkpoint to detect 20 car keypoints on annotated VeRi-776 \cite{vehicle-orientation-invariant-2017}.
We can observe that 94.5\% keypoints can be correctly predicted within 10 pixels (4\% of the image size) to the ground truth.
The domain shift between VeRi776 and Cars196 makes the predictions on the latter less accurate.
We visually inspect the predicted keypoints on Cars196 and conclude that many keypoints are detected correctly.
We use the predicted keypoints to train our model.

\subsection{Results}

\begin{table}[]
\begin{center}
\begin{tabular}{llll}
Method / Backbone model & CMC@1 & CMC@5 & mAP \\ \hline
AAVER RN50 \cite{vehicle-dual-path}    & 88.68  &  94.10  &  58.52 \\
AAVER RN101 \cite{vehicle-dual-path}   & 88.97  &  94.70  &  61.18 \\
VANet GL \cite{vehicle-viewpoint-aware} & 89.78 & 95.99 & 66.34 \\
BS M \cite{vehicle-triplet-baseline} & 90.23 & 96.42 & 67.55 \\
$\text{PAMTRI}^{*}$ DN121 \cite{vehicle-pamtri} & 90.94  &  96.72  &  65.44 \\
\hline
KAE-Net RN50 (ours) & \textbf{93.62} & \textbf{96.84} & \textbf{70.89}
\end{tabular}
\end{center}
\caption{Comparison with recent methods and state-of-the-art on VeRi-776.
(*) we show results on real data only without additional synthetic data.
Backbone networks: RN50 - ResNet50, RN101 - ResNet101, DN121 - DenseNet121, GL - GoogLeNet, M - Mobilenet.}
\label{tab:veri-results}
\end{table}

\begin{table}[]
\begin{center}
\begin{tabular}{l|cccc}
Method         & R@1  & R@2   & R@4    & R@8 \\ \hline
w/ KAE-Blocks  & \textbf{74.2}   & \textbf{83.3}   &  \textbf{89.1} &  \textbf{93.2} \\
w/o channel scaling  & 71.1 & 81.1  & 87.9 & 92.8 \\
w/o KAE-Blocks       & 68.8 & 79.2 & 86.9   & 92.4 \\
Baseline / ResNet50  & 66.9 & 77.4 & 85.3  & 91.1 \\
\end{tabular}
\end{center}
\caption{Ablation study on CUB-200-2011. We compare performance of the baseline model, KAE-Net without KAE-Blocks and KAE-Net with KAE-Blocks but without channel rescaling part.}
\label{tab:cub-baseline}
\end{table}

\begin{figure}[t]
\begin{center}
   \includegraphics[width=1\linewidth]{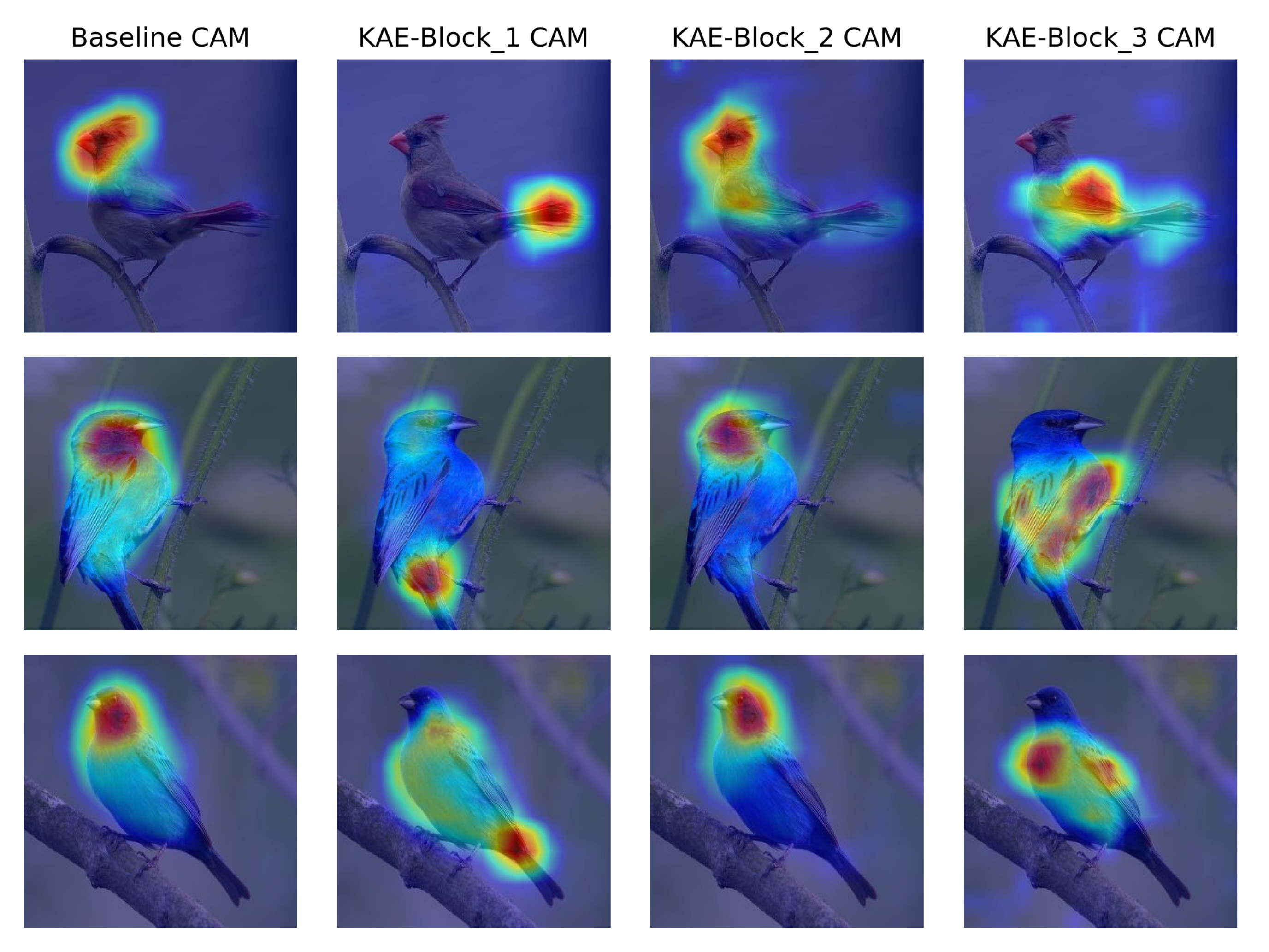}
\end{center}
   \caption{Visualization of class activation maps (CAMs) with Grad-CAM \cite{gradcam} of the baseline model and KAE-Blocks for different keypoints. The baseline model focuses mainly on the head while KAE-Blocks attend to various body parts. Each row shows CAMs for one image.}
\label{fig:cub_resnet_fmaps}
\end{figure}

\begin{figure}[t]
   \includegraphics[width=1\linewidth]{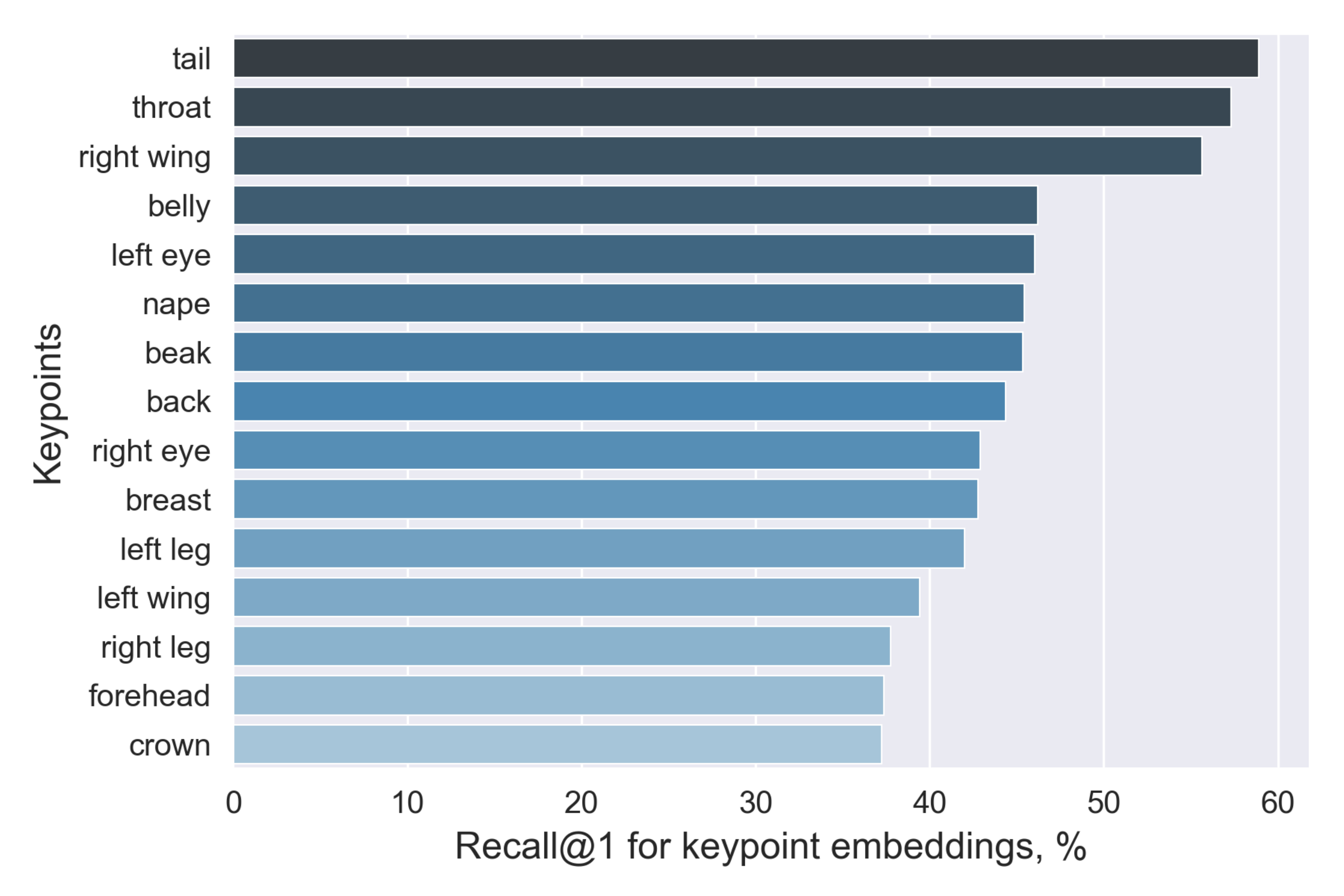}
   \caption{Recall@1 on subvectors of keypoint embeddings on CUB-200-2011. Subvectors of embeddings corresponding to a tail, a throat and a right wing have the highest recall.}
\label{fig:recall-kp-emb}
\end{figure}

\begin{figure*}[t]
\begin{center}
   \includegraphics[width=0.93\linewidth]{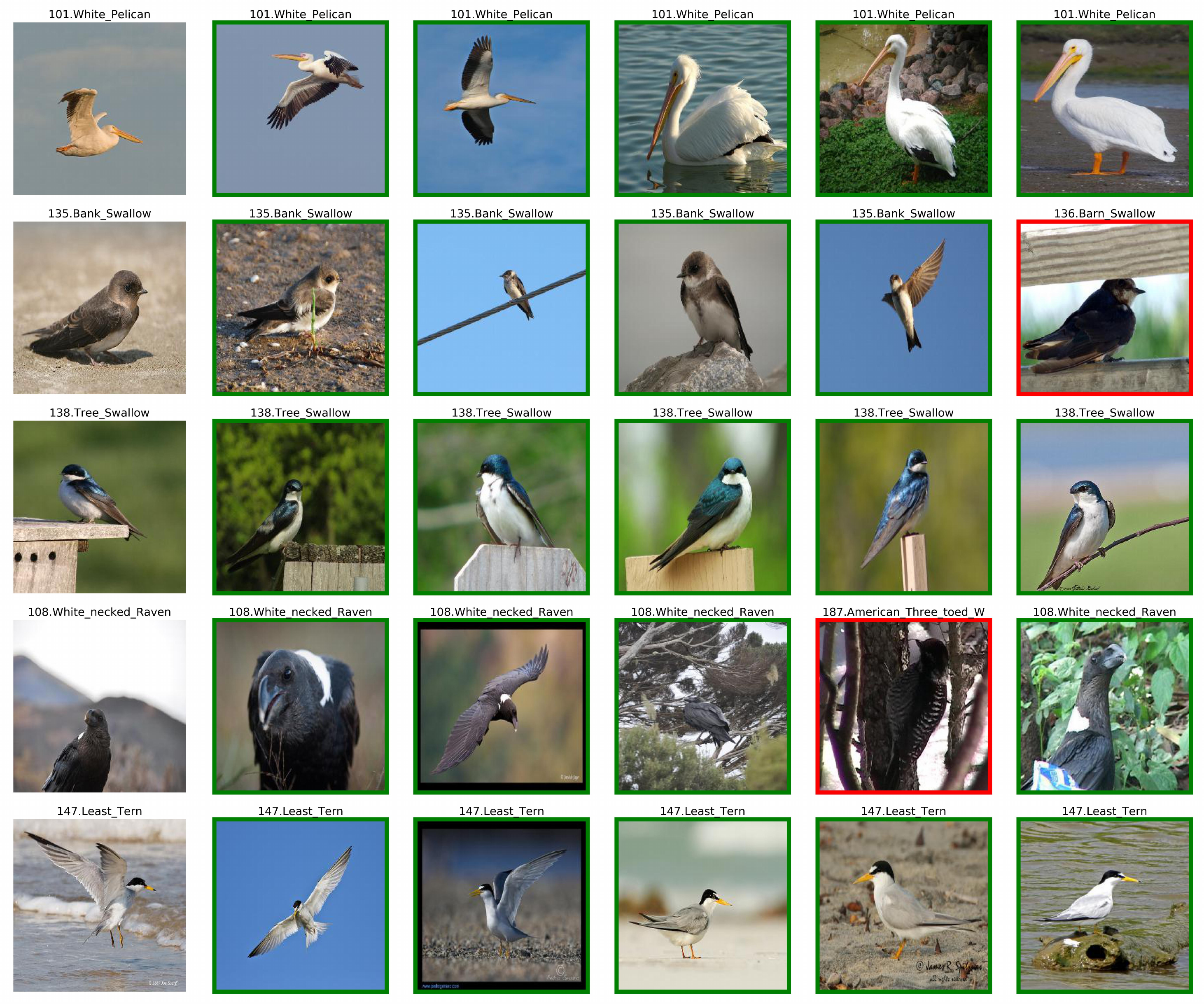}
\end{center}
   \caption{Qualitative visualization of the performance of KAE-Net on image retrieval on CUB-200-2011. The first image in each row is a query image. The top 5 matched test images are shown for each query image. Green and red boxes represent the same class (true) and different classes (false), respectively.   \textbf{Best viewed in colour.}}
\label{fig:cub-preds}
\end{figure*}

\begin{figure*}[t]
\begin{center}
   \includegraphics[width=0.93\linewidth]{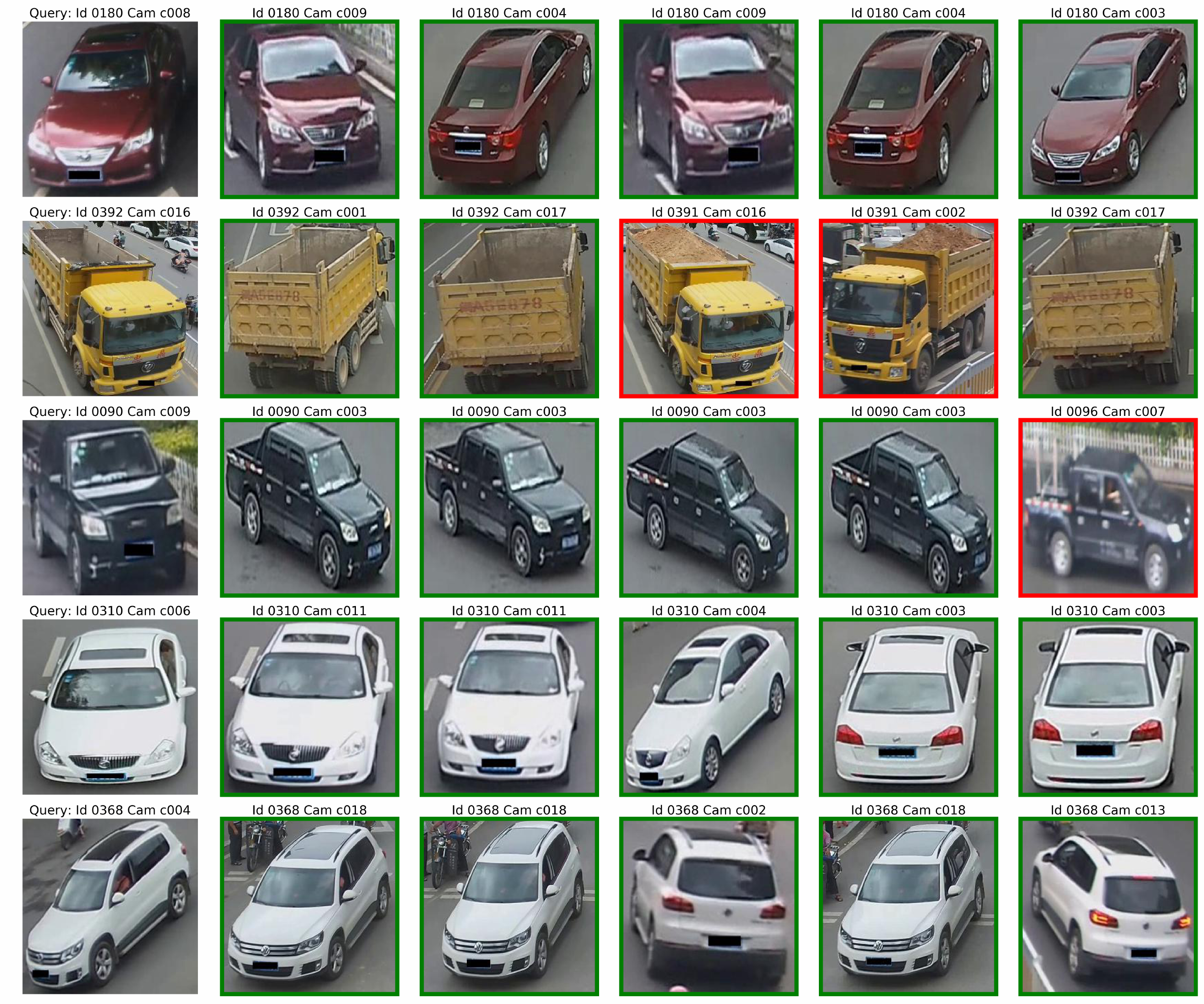}
\end{center}
   \caption{Qualitative visualization of the performance of KAE-Net on cross-camera vehicle re-identification on VeRi-776. The first image in each row is a query image. The top 5 matched gallery images are shown for each query image. Green and red boxes represent the same identity (true) and different identities (false), respectively.   \textbf{Best viewed in colour.}}
\label{fig:veri-preds}
\end{figure*}

The results of our KAE-Net against the state-of-the-art for image retrieval on CUB-200-2011 and Cars196 are reported in Table~\ref{tab:cub-cars-results}.
Most previous works in image retrieval are metric learning methods. 
Our method adds pose information to the embedding learning and outperforms the previous works by large margin achieving 74.2\% and 91.1\% Recall@1 on CUB-200-2011 and Cars196 respectively.

We further evaluate our method on VeRi-776 dataset, and compare it against the state-of-art methods of \textit{dual-path model with adaptive attention (AAVER)} \cite{vehicle-dual-path}, \textit{viewpoint aware network VANet} \cite{vehicle-viewpoint-aware}, \textit{pose-aware multi-task learning model (PAMTRI)} \cite{vehicle-pamtri}, as well as the strong baseline with the triplet loss~\cite{vehicle-triplet-baseline}.
All above methods use visual information from the image and do not use meta data such as car plates registration numbers.

We further compute the results without re-ranking and without usage of any external data (real or synthetic) as shown in Table~\ref{tab:veri-results}.
There is a variety of backbone models used in previous works.
Our method with the backbone ResNet50 outperforms previous methods and reaches 93.62\% CMC@1, 96.84\% CMC@5 and 70.89\% mAP.

\textbf{Ablation Study:} We perform ablation study on CUB-200-2011 dataset (Table~\ref{tab:cub-baseline}). The baseline network is obtained by fine-tuning convolutional layers of ResNet50 with the global average pooling layer and one fully connected layer on top to output an embedding of length 2048 (equal to the number of feature maps).
The baseline network is optimized with the triplet loss and reaches 66.9\% accuracy with the Recall@1.
Our baseline is strong and shows competitive performance with methods from Table~\ref{tab:cub-cars-results} because it uses large embeddings of length 2048. 

As shown in Table~\ref{tab:cub-baseline}, removing the channel scaling part from KAE-Blocks reduces Recall@1 from 74.2\% to 71.1\%.
We further remove KAE-Blocks from the model to evaluate its contribution to the performance. 
Removal of KAE-Blocks drops the retrieval Recall@1 to 68.8\% compared to 74.2\% of KAE-Net with KAE-Blocks. As observed, KAE-Net without KAE-Blocks reconstructs heatmaps and learn embeddings from the same feature map $F_C$.

\textbf{Visualization:} 
We analyse our proposed method by visualizing class activation maps (CAMs) of the predictions using Grad-CAM \cite{gradcam} on three images of birds (Figure~\ref{fig:cub_resnet_fmaps}).
We observe that CAMs of the baseline model are activated mainly around the head while KAE-Blocks focus on different parts of the bird guided by keypoints.

We further analyse the segments $h_i$ of keypoint aligned embeddings $h$ learnt with KAE-Net on CUB-200-2011.
An image embedding consists of 15 subvectors $h_i$ (for 15 keypoints on CUB-200-2011) with the length of 64 each.
We perform the evaluation procedure using each subvector of an embedding separately to investigate which keypoints are the most discriminative.
Figure~\ref{fig:recall-kp-emb} shows that areas around tail, throat and right wing have the highest recall.

Lastly, we show some examples of matches using our proposed method on CUB-200-2011 in Figure~\ref{fig:cub-preds} and VeRi-776 in Figure~\ref{fig:veri-preds}. 
Our model found the matching bird classes with the presence of large pose variations including flying and sitting poses (Figure~\ref{fig:cub-preds}). 
As shown in Figure~\ref{fig:veri-preds}, our approach found the same vehicle in different cameras even when the viewpoint has changed from the front to the back view.


\section{Conclusion}
\label{sec:conclusion}
In this work, we propose a multi-task method to learn pose invariant embeddings.
The method is generic and does not use specific domain information. In future,  we will investigate if additional domain knowledge such as using symmetry of some body parts for birds (e.g.~eyes, wings and legs) or combining some keypoints for cars can further improve the performance of our approach. We will further examine relaxing the requirement of ground truth keypoint annotations by incorporating a pose estimation network into the re-identification pipeline.


{\small
\bibliographystyle{ieee_fullname}
\bibliography{egbib}
}

\end{document}